\documentclass[conference]{IEEEtran}
\IEEEoverridecommandlockouts
\usepackage{cite}
\usepackage{amsmath,amssymb,amsfonts}
\usepackage{algorithmic}
\usepackage{graphicx}
\usepackage{textcomp}
\usepackage{xcolor}
\usepackage{moreverb,url}
\usepackage[]{hyperref}
\usepackage{dirtytalk}
\usepackage{graphicx} 
\usepackage{subfigure}
\usepackage{wrapfig}
\usepackage{url}
\usepackage{textcase}
\def\BibTeX{{\rm B\kern-.05em{\sc i\kern-.025em b}\kern-.08em
    T\kern-.1667em\lower.7ex\hbox{E}\kern-.125emX}}
\begin{document}

\title{A Hierarchical Variable Autonomy Mixed-Initiative Framework for Human-Robot Teaming in Mobile Robotics\\
{\thanks{\textsuperscript{*} First and second authors contributed equally.}}
\thanks{This work was supported by the EPSRC grant EP/P01366X/1.}
}
\author{\IEEEauthorblockN{1\textsuperscript{st} Dimitris Panagopoulos*}
\IEEEauthorblockA{\textit{Industrial Design} \\ {\textit{\& Production Engineering}} \\
\textit{University of West Attica}\\
Athens, Greece \\
dimipan.auto@gmail.com}
\and
\IEEEauthorblockN{2\textsuperscript{nd} Giannis Petousakis*}
\IEEEauthorblockA{\textit{Cognitive Robotics Lab (COROLAB)} \\
\textit{University of Manchester}\\
Manchester, United Kingdom \\
ioannis.petousakis@postgrad.manchester.ac.uk}
\and
\IEEEauthorblockN{3\textsuperscript{rd} Aniketh Ramesh}
\IEEEauthorblockA{\textit{Extreme Robotics Lab (ERL)} \\
\textit{University of Birmingham}\\
Birmingham, United Kingdom \\
AXR1050@student.bham.ac.uk}
\and
\IEEEauthorblockN{4\textsuperscript{th} Tianshu Ruan}
\IEEEauthorblockA{\textit{Extreme Robotics Lab (ERL)} \\
\textit{University of Birmingham}\\
Birmingham, United Kingdom \\
txr094@student.bham.ac.uk}
\and
\IEEEauthorblockN{5\textsuperscript{th} Grigoris Nikolaou}
\IEEEauthorblockA{\textit{Industrial Design} \\ {\textit{\& Production Engineering}} \\
\textit{University of West Attica}\\
Athens, Greece \\
nikolaou@uniwa.gr}
\and
\IEEEauthorblockN{6\textsuperscript{th} Rustam Stolkin}
\IEEEauthorblockA{\textit{Extreme Robotics Lab (ERL)} \\
\textit{University of Birmingham}\\
Birmingham, United Kingdom \\
r.stolkin@bham.ac.uk}
\and
\IEEEauthorblockN{7\textsuperscript{th} Manolis Chiou}
\IEEEauthorblockA{\textit{Extreme Robotics Lab (ERL)} \\
\textit{University of Birmingham}\\
Birmingham, United Kingdom \\
m.chiou@bham.ac.uk}
}

\maketitle

\begin{abstract}

This paper presents a Mixed-Initiative (MI) framework for addressing the problem of control authority transfer between a remote human operator and an AI agent when cooperatively controlling a mobile robot. Our Hierarchical Expert-guided Mixed-Initiative Control Switcher (HierEMICS) leverages information on the human operator's state and intent. The control switching policies are based on a criticality hierarchy. An experimental evaluation was conducted in a high-fidelity simulated disaster response and remote inspection scenario, comparing HierEMICS with a state-of-the-art Expert-guided Mixed-Initiative Control Switcher (EMICS) in the context of mobile robot navigation. Results suggest that HierEMICS reduces conflicts for control between the human and the AI agent, which is a fundamental challenge in both the MI control paradigm and also in the related shared control paradigm. Additionally, we provide statistically significant evidence of improved, navigational safety (i.e., fewer collisions), LOA switching efficiency, and conflict for control reduction.
\end{abstract}


\section{Introduction}

In Human-Robot Teaming (HRT), the Human-Robot Interaction (HRI) should facilitate mutual support between robots and humans towards achieving a common goal, and allow intervention by either agent (i.e., humans and robots) when necessary. This is especially true for high-risk safety-critical applications such as robot-assisted disaster response scenarios \cite{Hong2018}, remote inspection and manipulation, or exploration, in which human operators control remotely situated robots. Cooperation between human and AI agents (e.g., robot's AI agent) can provide essential help in preventing errors and allow both agents to establish a complementary relationship that achieves greater system performance and effective task completion. Such systems, which keep the human in the control loop, could benefit from a coordinated transfer of control between agents (i.e., human and AI agent) to safely operate the mobile robot \cite{9146977}. This is especially true for variable autonomy robotic systems, such as shared control \cite{pappas, Selvaggio2021, Abbink2018} or Mixed-Initiative (MI) \cite{chiouMI} systems that can adjust their autonomy level in response to changes in the environment.

In MI systems humans and AI agents are given equal authority to initiate Levels of Autonomy (LOA) switching (e.g., between autonomy and teleoperation), seizing or relinquishing the control of the mobile robot \cite{Jiang2015}. Previous work \cite{chiouMI} identified conflict for control between the human operator and AI agent during implicit hand-off coordination as a major challenge for the MI field. Conflict for control refers to a situation in which the human operator and the robot’s AI aggressively attempt to take control of the robot, causing a cycle of the two agents overriding each other's commands. For instance, commands issued by the human operator (e.g., switching to teleoperation to explore an area of interest) can be misinterpreted by the robot and inferred as performance degradation. Generally, conflict happens because the AI agent lacks understanding of higher-level of abstraction information about the environment and their human teammate (e.g., human state or intent, scene understanding and other semantics) \cite{chiouMI, Nemics}.

According to \cite{flemisch}, it is recommended that the transfer of control authority has to be adaptable to the dynamic states of both agents. Chiou et al. \cite{fieldExercise}, among others, found that the main degradation in performance comes from the cognitive and attentional demands on the operator. This finding suggests that emphasis should be given to control authority transfer informed on the human partner's cognitive state.

This paper proposes a framework that enables the robot's AI agent to collect and use information about the cognitive state and intent of their human teammate in the context of mobile robot navigation tasks. Perceiving information about the human operator is introduced through various sensor modules, and functionalities. These capabilities are stratified based on their criticality (i.e., the importance of executing a task correctly as evaluated based on its negative effect in case of failure or error \cite{Yanco2002ATF}) as conferred by expert knowledge on the ability of the Human-Robot system (HRS) to perform effectively.

The contribution of this paper is twofold. First, a Hierarchical Mixed-Initiative Control Switcher (HierEMICS) framework is proposed that leverages implicit inference of human state and human intent and arranges its priorities based on the criticality of each case. Second, we present a practical realization of the proposed framework based on the cognitive availability \cite{Petousakis2020} and intent recognition \cite{BOIR} of the human operator. Additionally, we systematically evaluate the framework realization in a high-fidelity robot simulation, in a disaster response robotics exploration and navigation scenario, and compare it to a state-of-the-art Expert-guided Mixed-Initiative Control Switcher (EMICS) \cite{chiouMI}.

\section{Related Work}

In HRI, Scholtz \cite{1174284} identifies various roles for the humans such as supervisor, operator, teammate, mechanic/programmer, and bystander. In our research, the human acts as an operator and a peer/teammate to the AI agent, that in turn can actively assist the human with the task and control transfer allocation.

In this context, conflict for control has emerged in some related works \cite{medina, Ludwig, chiouMI}, which include shared control in manipulation tasks, assistive robotics and autonomous driving. According to \cite{norman}, achieving robust shared control interactions between humans and automated systems requires humans to be actively involved and adequately informed, while understanding each other's intents correctly. Such systems are usually haptic shared control (HSC) \cite{Abbink}, in which a human operator and an AI are allowed to exert their respective forces on an operational interface in order to control a vehicle. Similarly, due to the driver's time-varying behaviour, De Jonge et. al \cite{DeJonge2016} reduced conflict, between human operators and a haptic shared controller, by modifying supported trajectories in a trial-by-trial adaptation. However, the MI systems that are focused on discrete transfer of control authority (i.e., LOA switching), do not account for such solutions. Both agents in discrete LOAs MI systems have the authority to initiate or completely override each other’s commands \cite{chiouMI}. Owan et al. \cite{Owan2017} created an explicit consensus integrated scheme that completes a control trade only with the consent of the agent that did not propose the trade. Chiou et al. \cite{Nemics} leveraged negotiation theory to overcome conflicts for control by having the AI and the human negotiate the transfer of control authority between each other under a time-based concession strategy scheme.

Most relevant to our paper, in the sense that they consider the operator's intent and attention in their policies, are the following works. Li et al. \cite{Li} have implemented a fuzzy logic shared control method for calculating a continuous shared control coefficient based on the driver's intention and the risk of collision. In \cite{Gold2013}, during the transfer of control, the behaviour of the driver is observed, assessing whether they are attention-enabled to take over the driving task in a safe manner.

However, the works found in the literature propose that actions initiated by AI are often focused on adjusting the level of assistance, reacting to sensor inputs (e.g., obstacles), predefined priorities or haptic feedback rather than providing explicit policies for avoiding conflicts. This paper aims to address the gaps by introducing an MI control system that can address and reduce conflict for control. This is achieved through the implementation and, criticality-based stratification of new information about the human partner. Since criticality is a highly subjective measure \cite{Yanco2004}, we provide a structured way to take into account factors and aspects while relating them in view of its practical integration with human experts' prior knowledge.

\section{Hierarchical Mixed-Initiative Level-of-Autonomy Switcher}

\subsection{Hierarchical framework}

\label{HierEMICS}

One of the fundamental motivations of MI control schemes is improving the HRS performance. HRS are used in a wide variety of applications and make use of a large number of variables. In the effort to improve these systems, even more diverse modules are often introduced. Therefore, it is necessary to introduce new capabilities in a structured way, in our case by using a hierarchical framework. 

\begin{figure}
	\centering
	\includegraphics[width=0.99\columnwidth]{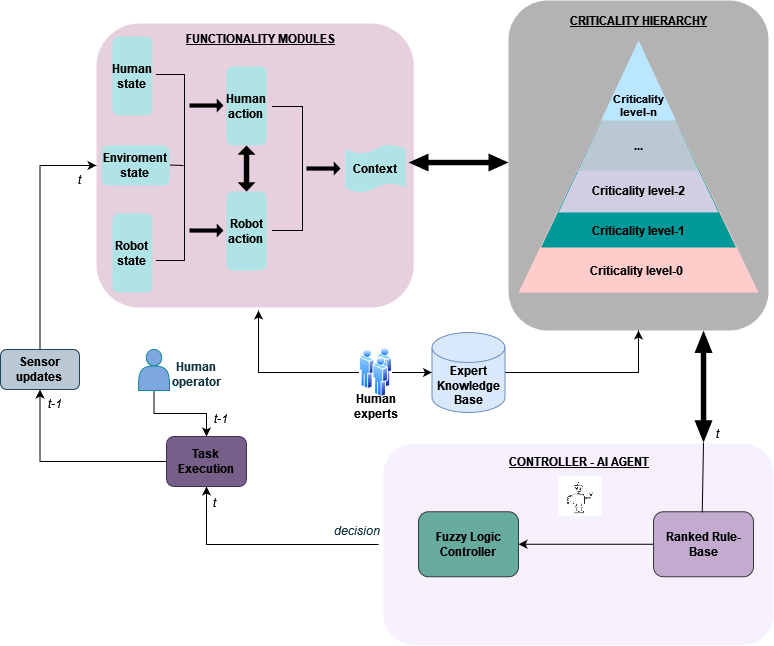}
	\caption{Representation of HierEMICS framework with the three nodes shown.}
	\label{fig:system}
\end{figure}

As shown in Fig. \ref{fig:system}, our main approach entails creating a three-node universal framework that allows robots to be equipped with an array of sensors and functionalities. This enables them to collate information on the human operator, the robot, and the environment and stratify them based on how critical they are to the overall system's performance. Within this framework, we assume that criticality is obtained by using expert knowledge to assess the impact each factor has on the ability of the HRS to function and perform its assigned task. The relationship between the different nodes of the framework is bidirectional, meaning we can infer the criticality of a task based on the available modules or choose to introduce certain modules to achieve a known critical task.

The three nodes of the framework, are Functionality Modules, Criticality Hierarchy, and the resulting Controller-AI Agent.

\subsubsection{Functionality Modules}
This node contains all information available in the context of the system that is being designed. The information provided in this section can be low level information about the state of humans, robots and the environment, usually through using new sensors or already existing data in novel ways. Additionally, this node can include high level information about the actions of humans and robots and can result in higher level of abstraction context being deduced from those actions. In our hierarchical controller, we exploit the following two types of information about the human operator to better inform the AI decisions when transferring control between team members.

\subsubsection{Criticality Hierarchy}
The second node of the framework takes the various pieces of information from the previous node, and is focused on arranging them and their combinations based on their criticality. The modular nature of the framework allows experts from different fields to contribute to the creation of the criticality hierarchy. To assist in this effort, an Analytic Hierarchy Process (AHP) \cite{SAATY1987161} is applied by the expert to stratify the utility that can be extracted from the various modules and functionalities. The resulting hierarchy is then used to compile the rule-base for the controller.

\subsubsection{Controller - AI Agent}

The final stage of the framework serves as a mapping from information to action that depends on the established hierarchy. Similar to \cite{Petousakis2020}, a hierarchical fuzzy approach has been adopted, that prioritizes the first rule to be activated, over the others.

\subsection{HierEMICS Implementation}

The framework proposed in this paper is extending upon the MI framework (EMICS) created by Chiou et al. \cite{chiouMI}. In this context, MI system refers to an expert-guided system where both the human operator and the AI agent have the authority to intervene, initiate actions, and coordinate the shift of initiative \cite{Jiang2015}.

Regarding EMICS, there are two assumptions made. First, the human operator is willing to both relinquish control to and receive control from the AI agent. Second, the agent (i.e., either the human or AI agent) to which the control may be handed over is capable of correcting the performance degradation as expressed by the error.

According to these assumptions, in EMICS, the control is transferred (e.g., LOA switching) to the agent currently being most competent towards the completion of the task. This is achieved by comparing the current performance and the expected optimal performance yielding an online task effectiveness metric called goal-directed motion error. Essentially, this metric expresses how effectively the system performs the navigation task based on the difference between the robot's current motion (i.e., linear velocity) and the motion required of the robot to achieve its goal (i.e., reach a target location) according to an expert planner \footnote[2]{Navigation planner default plug-in in Robot Operating System (ROS) uses the Dijkstra algorithm and the dynamic window approach.}. The parameters of EMICS were tuned based on human-initiative LOA switching data from previous experiments.

In our proposed HierEMICS system there are two objectives to be achieved: (1) facilitating efficient LOA switching by both the human operator and the AI agent; and (2) reducing conflict for control. To this end, the AI agent is provided with the ability to monitor the human operator's cognitive availability and predict their navigational intent. These two modules were chosen as a way to evaluate the effectiveness of the proposed framework, leaving beyond the current scope of this work the incorporation of modules that would provide information about the robot \cite{Vitals2022} or the environment \cite{RAN2021389}.

\textbf{Cognitive availability}. Human operators in SAR scenarios are likely to engage in non-driving related tasks, regardless of whether the robot is in teleoperation or autonomy driving mode. Due to their involvement in these tasks for a prolonged time \cite{ZEEB201765} or due to the stress or physical fatigue they encounter \cite{8569545}, a longer time might be required to shift their attention back to driving. Correctly detecting the operator's cognitive availability offers an opportunity to make a practical decision of control transfer in real time, thus improving mobile robot safety. Cognitive availability indicates whether the operator is available to focus on controlling the mobile robot or not. In particular, the cognitive availability module uses the Deepgaze library by Patacchiola and Cangelosi \cite{PATACCHIOLA2017132} allowing the system to capture the operator's head image, and extract their head angles and head pose using deep learning. This method of estimating the cognitive availability of the operator, to inform the AI agent on the status of the human operator, has been experimentally evaluated in \cite{Petousakis2020}. Further evaluation of the method is outside of the current paper's scope. 

\textbf{Intent recognition}. In addition to using \textit{cognitive availability} that determines the attention of the human operators, it is also essential to enable the robot to have a knowledge of their actions and what it is they are trying to accomplish (e.g., explore an area of interest). A probabilistic approach is adopted that can recognize the human operator's intended navigational goal. Specifically, we exploit the work made by Panagopoulos et al., in \cite{BOIR}, in which a Bayesian Operator Intent Recognition (BOIR) algorithm was proposed and experimentally validated in multiple scenarios. The algorithm enables the AI agent to assign probabilities by taking into account the path distance and the angle of the robot for each possible navigational goal.

\begin{figure}
	\centering
	\includegraphics[width=0.99\columnwidth]{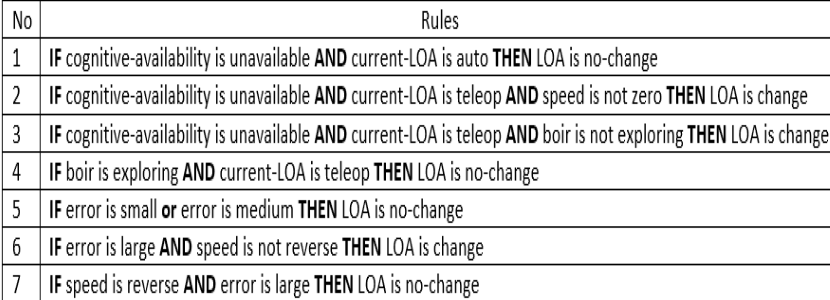}
	\caption{Fuzzy Rule Base}
	\label{fig:rulebase}
\end{figure}

Our previous work has already established the efficacy of the cognitive availability and navigational intent recognition modules. The main focus of this paper, thereby, is on the creation and evaluation of the proposed HierEMICS framework that incorporates the aforementioned modules. In HierEMICS's ranked rule base (see Fig. \ref{fig:rulebase}), the most critical level focuses on ensuring the safe operation of the system. To that end, the cognitive availability of the operator is used to assure they provide driving input only when they are attending the Operator Control Unit (OCU) screen. The next level supports the reduction of conflict for control and assisting the operator in their effort. Using the navigational intent recognition method (BOIR) described above, HierEMICS is able to monitor when the operator is engaging with a meaningful goal. The operator's driving input is monitored in real time, by the AI agent, along with their inferred navigational intent. This information and the prior knowledge of the area allow the AI agent to discern whether the human is \say{\textit{exploring}}, or not, an object of interest, adjusting the LOA switching policies in accordance with each case.

\section{Experimental Evaluation} \label{experiments}

An experiment was conducted to verify the following hypothesis: The conflict for control and its negative effects (e.g., collisions) can be reduced via the use of HierEMICS and its higher level information about the human operator when compared to EMICS \cite{chiouMI}. Specifically, we set out to a) evaluate the functionality of the proposed HierEMICS, b) identify the impact of HierEMICS on the HRS performance (i.e., conflict for control, collisions, LOA switching, workload), and c) investigate the effect of a possible reduction in conflict for control, to the perceived mental workload of the operator.

\subsection{Experimental Setup \& Participants}

\begin{figure}
	\centerline{\subfigure[]{\includegraphics[width=0.5\columnwidth]{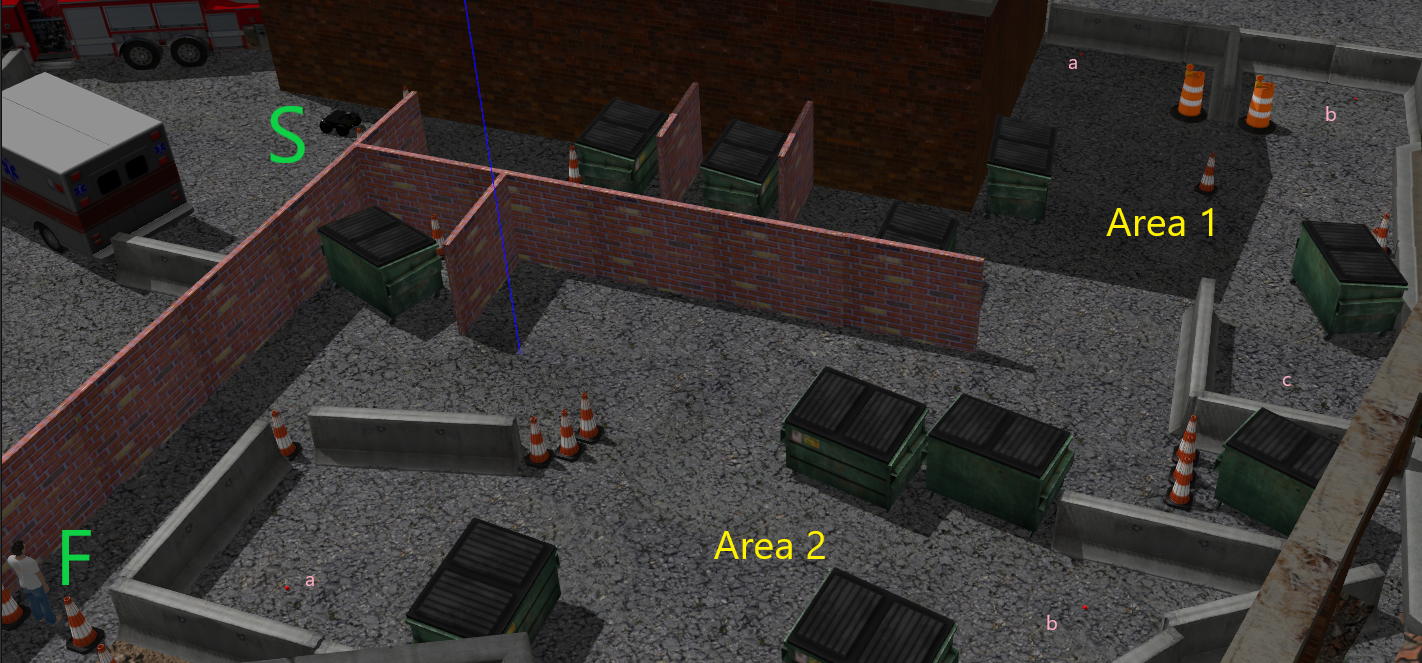}
			\label{fig:arena}}
		\hfil
		\subfigure[]{\includegraphics[width=0.48\columnwidth]{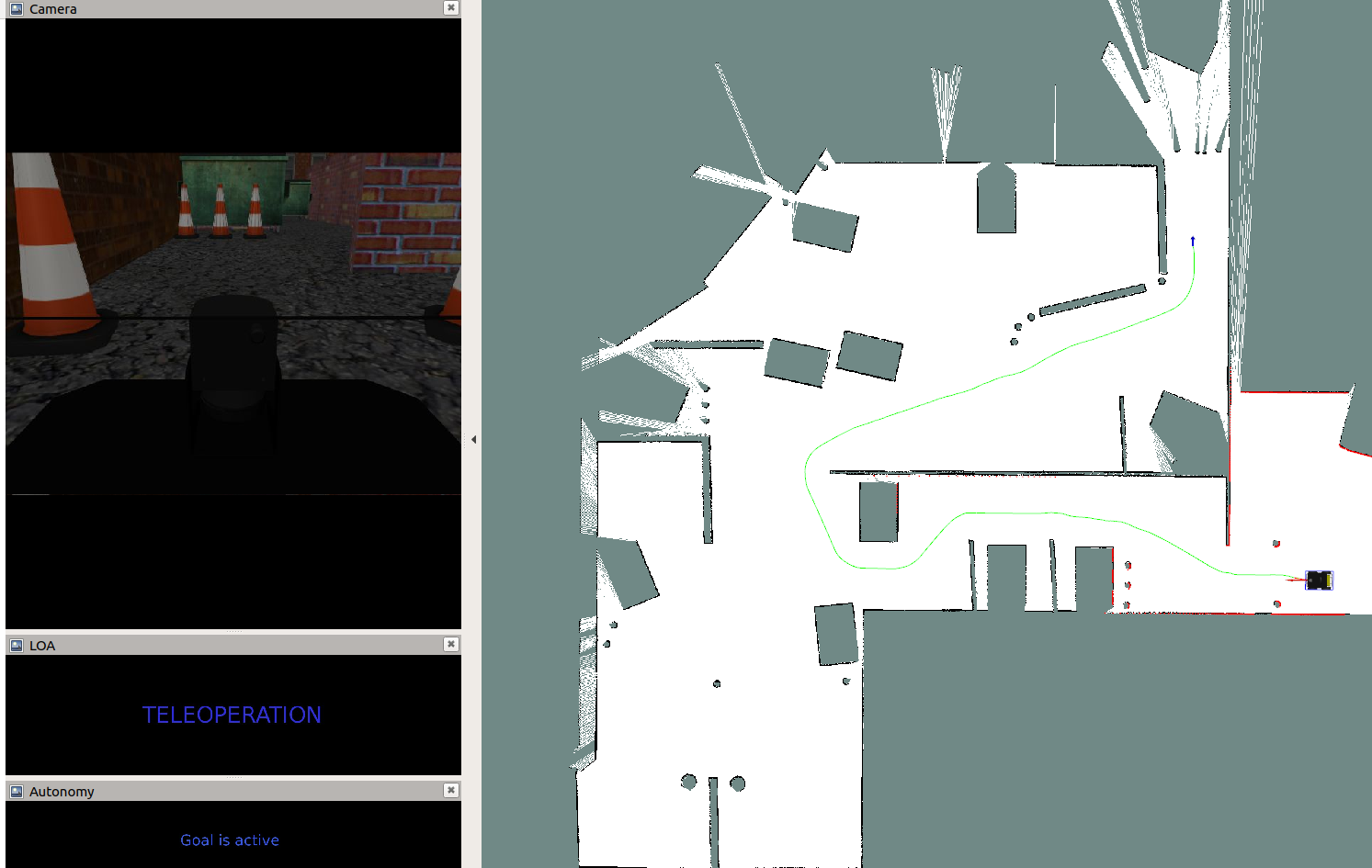}
			\label{fig:gui}}}
	\caption{\textbf{\ref{fig:arena}:} The arena used in the experimental evaluation, simulating a disaster response scenario. \textbf{\ref{fig:gui}:} The Graphical User Interface (GUI). \textbf{Left:} Video feed from the camera, the control mode in use and the state of the robot. \textbf{Right:} The map (as created by SLAM) showing the position of the robot, the current goal (blue arrow), the AI planned path (green line).}
	\label{fig:arenagui}
\end{figure}

Our experimental setup is comprised of a simulated mobile robot task and an OCU that enables interaction between a human operator and the simulated mobile robot. The OCU consists of a mouse, a joystick and an off-the-shelf webcam as input devices, a monitor screen displaying the Graphical User Interface (GUI) (see Fig. \ref{fig:arenagui}) and a laptop running the simulated environment (see Fig. \ref{fig:arena}) and the corresponding software. The simulation was based on a high-fidelity robotics simulation system with a realistic physics engine, named Gazebo. The software and related functionalities were developed in the Robot Operating System (ROS). The mobile robot was equipped with a laser range finder and an RGB camera. It was, also, capable of operating in two different LOAs (i.e., control modes): teleoperation (human operator fully in control of navigation via joystick) and autonomy (AI agent has full control over its navigation).

The subjects participating in this experiment were 11 males and 2 females with a mean age of $M = 27$ $(SD = 1.41)$. To ensure a minimum skill level and system understanding, each participant undertook extensive standardized training \cite{effectsChiou}.

\subsection{Experimental Procedure}
We evaluated the proposed HierEMICS framework, against the EMICS in a navigation and exploration scenario. The participants were assigned to complete the scenario twice, one trial for each controller (i.e., EMICS and HierEMICS). The two controllers' order was counterbalanced among the participants to account for learning effects. 

In each of these trials, the operators were tasked with navigating a realistic simulated mobile robot from point S (Start) to point F (Final) in the arena by means of two LOAs (autonomy \& teleoperation). The HRS had to navigate and explore a simulated disaster site that contained two Areas-of-Interest (AOI). Within these areas, the human operators were advised to search and identify various points-of-interest (POI). In the first AOI, there were three POIs, while in the second AOI, the number of POIs to be explored was two. Once the required POI were investigated, the operator advanced to the next AOI and eventually the final location of the arena. 

During the experiment, a human operator performance degradation factor (i.e., secondary task) was introduced, in the form of mental rotations of 3D objects \cite{Ganis2015}, simulating the varied workload operators face during a real disaster response scenario. Similarly to previous experiments \cite{Petousakis2020}, the goal was to temporarily increase the mental workload of the participant. The workload was measured using NASA Task Load Index (NASA-TLX) questionnaire \cite{Hart.2006} which was administered to the participant after each trial. While filling out the questionnaire for the second trial, the participants were permitted to view and modify their answers to the first to allow them to adjust the relative evaluation of the two systems tested (i.e., EMICS and HierEMICS). Lastly, two ways were adopted to record conflicts for control, 1) the experimenters noted instances when they believed conflict for control appeared and 2) we had the participants verbally state when they felt there was a conflict.

\section{Measures \& Results} \label{results}
To evaluate the effectiveness of the proposed framework, a number of objective and subjective measurements were recorded and analyzed. For the data that conformed to a normal distribution based on Kolmogorov–Smirnov test, paired sample t-tests were used, while for the rest the non-parametric Wilcoxon signed-rank test was applied. We consider results statistically significant if $p<.05$.

\textbf{Time-to-completion:} The overall task completion time was averaged across trials for each controller. Statistical analysis showed no statistically significant difference (\textit{$t = -1.053, p > .05$}) between EMICS ($M = 283s, SD = 32s$) and HierEMICS ($M = 272s, SD = 29s$). 

\textbf{Number-of-collisions:} The participants completed the navigation task with statistically significant (\textit{$Z = -2.266, p = .023$}) less collisions, when using HierEMICS ($M = 0.23, SD = 0.44$) compared to EMICS ($M = 1.23, SD = 1.24$). It is worth mentioning that while using HierEMICS, 10 out of 13 participants completely avoided collisions during task execution. On the contrary, while using EMICS, the number of participants that avoided collisions was reduced to 5.     

\textbf{Conflicts-for-control:} The experimenter's data indicated that there were statistically significant (\textit{$Z = -3.083, p = .002$}) more conflicts with EMICS ($M = 9.92, SD = 4.97$), compared to HierEMICS ($M = 4.77, SD = 3.56$). Similarly, the participant's reported conflicts showed statistically significant (\textit{$Z = -3.186, p = .001$}) more conflicts with EMICS ($M = 9.00, SD = 5.12$), compared to HierEMICS ($M = 3.85, SD = 3.02$).

\textbf{Number-of-LOA-switches:} We measured the number of \textit{AI initiated switches}, and the total number of switches made by both agents\textit{total switches}. Both of these measurements show a statistically significant reduction of AI initiated switches (\textit{$Z = -3.063, p = .002$}), and total switches (\textit{$Z = -2.975, p = .003$}). On average, EMICS produced ($M = 11.62, SD = 4.94$) AI initiated switches, and ($M = 28.15, SD = 9.98$) total LOA switches while HierEMICS produced ($M = 5.54, SD = 4.31$) AI initiated, and ($M = 16.38, SD = 9.80$) total switches.

\textbf{Workload-measurement with NASA-TLX:} Participant's workload while using EMICS ($M = 43.66, SD = 20.37$) compared to using HierEMICS ($M = 40.00, SD = 16.33$) did not show any statistical significance (\textit{$t = -1.558, p > .05$}). There was a trend of improved performance in the dimensions of effort and frustration between EMICS and HierEMICS. With the effort showing (\textit{$t = -2.132, p = .054$}) between EMICS ($M = 51.54, SD = 25.61$) and HierEMICS ($M = 45.78, SD = 25.65$), and frustration showing (\textit{$t = -2.044, p = .063$}) between EMICS ($M = 39.23, SD = 30.81$) and HierEMICS ($M = 27.69, SD = 20.17$).

\section{Discussion} \label{discussion}

The experimental results confirm our hypothesis that HierEMICS, compared with EMICS, reduces conflict for control in mobile robot navigation tasks and as a result improves safety and LOA switching efficiency.

Specifically, both measurements of conflict for control (i.e., the one measured by experimenters and the one reported by the participants) agree on the a) value range; and b) significant reduction of conflict for control. This evidence suggests that introducing information on the human operator's intent to the system can lead to decreased conflicts for control as it improves the LOA switching policies. This improvement of the LOA switching policies is possible because HierEMICS can use the operator's intent to infer whether they are meaningfully exploring. The reduction observed (statistically significant) in the number of LOA switching, along with the similar time to complete and NASA-TLX scores, indicates HierEMICS also provided more efficient LOA switching.

Additionally, the results demonstrate statistically significant reduction in the number of collisions and hence improved navigation safety. This improvement can be ascribed to the reduction of conflicts for control. In our experiment when conflict for control occurred, especially in space constrained environment, collisions were frequent. This is attributed to operators attempting to maintain control, navigate the area, and avoid obstacles while encountering the insistent LOA switching initiated by the AI agent. When using HierEMICS, only 3 participants experienced a single collision each. In contrast, when using EMICS, 8 participants experienced collisions and anecdotally claimed that conflict was the main cause. This aligns with insights from previous studies that showed that most collisions occurred due to conflict for control \cite{Nemics}.

\section{Limitations and Future Work}

As a future direction HierEMICS can be extended by introducing more functionalities (e.g., multi-robot state monitoring \cite{Vitals2022}, operator's driving behaviour \cite{fessonia}, environment state \cite{RAN2021389}) and implemented into other robotic platforms. An implementation focusing on the robot's own state perception could be also included. By introducing a metric for monitoring the emergence of conflict for control, the robot could monitor its own state and adjust its policies accordingly. 

A limitation of the current study, which is also often seen in related work when researching such systems in an academic setting, is that the participant pool differs greatly in skill set and experience. This effect is even greater when trying to examine results with high variance in the measured metrics, which can be only partially mitigated with a systematic design. As a result, an important direction for future research is the evaluation of HierEMICS and related systems with expert participants. When such systems are deployed in the field, the human operator is usually an expert or at least has received extensive training in operating the platform and efficiently utilizing its capabilities. It would be worthwhile comparing the performance and also the feedback of expert operators to that of the normal university pool participants.

\section{Conclusion} \label{conclusion}

In this work, HierEMICS was presented, a framework that addresses the transfer of control authority between a human operator and the AI agent, when remotely controlling a mobile robot in navigation and exploration tasks. HierEMICS was created using a framework that streamlines the process of introducing new modules and sensors to the system in a structured way. Modules that provide information about the operator's intent and cognitive availability were used to adjust the AI agent's policies. An experimental evaluation of the created AI agent was conducted, inspired by a disaster response scenario, demonstrating that HierEMICS can reduce instances of conflict for control, collisions, and LOA switching as compared to the state-of-the-art EMICS.

\bibliography{thebibliography}
\bibliographystyle{IEEEtran}

\end{document}